\documentclass{article}

\usepackage[accepted]{icml2026}

\usepackage{amsmath}
\usepackage{amssymb}
\usepackage{booktabs}
\usepackage{graphicx}
\usepackage{hyperref}
\usepackage{url}
\usepackage{multirow}
\usepackage{xcolor}
\usepackage{microtype}

\icmltitlerunning{Accelerometry Biomarkers: A Tabular Health Benchmark}

\begin{document}

\twocolumn[
\icmltitle{Accelerometry-Derived Digital Biomarkers for Cardiometabolic Risk:\\
A Population-Representative Tabular Benchmark\\
with Uncertainty Quantification}

\icmlsetsymbol{equal}{*}

\begin{icmlauthorlist}
\icmlauthor{Federico Felizzi}{siiam,engene}
\end{icmlauthorlist}
\icmlaffiliation{siiam}{SIIAM, Rome, Italy}
\icmlaffiliation{engene}{enGene Statistics, Basel, Switzerland}
\icmlcorrespondingauthor{Federico Felizzi}{federico.felizzi@gmail.com}

\icmlkeywords{tabular data, digital biomarkers, conformal prediction,
              NHANES, cardiometabolic risk, uncertainty quantification,
              tabular foundation models}

\vskip 0.3in
]

\printAffiliationsAndNotice{}

\begin{abstract}

Structured tabular data is the dominant format in clinical medicine, yet
existing tabular benchmarks fail to reflect key properties of real-world
health data: complex survey sampling, demographic oversampling, clinically
validated outcomes, and fairness requirements across population subgroups.
We introduce the \emph{NHANES Accelerometry Cardiometabolic Benchmark},
derived from the National Health and Nutrition Examination Survey (NHANES)
2003--2006, comprising 1,381 adults with hip-worn accelerometry, fasting
laboratory biomarkers, dietary intake, and anthropometric measurements.
We evaluate three methods spanning the specialised-to-general spectrum of
tabular learning---ridge regression, XGBoost, and the tabular foundation
model TabPFN~v2---on prediction of glycated haemoglobin (HbA1c), fasting
triglycerides, and C-reactive protein (CRP) from accelerometry-derived
activity phenotypes and lifestyle covariates.
TabPFN~v2 achieves the best overall performance
(HbA1c R$^2$=0.156, CRP R$^2$=0.383), while triglycerides
remain largely unpredictable from lifestyle features alone
(R$^2 < 0.05$ across all models), consistent with known genetic dominance.
We apply \emph{split conformal prediction} to all models, producing
distribution-free 90\% prediction intervals, and evaluate demographic
coverage equity across sex and race/ethnicity subgroups.
Marginal coverage is close to the 90\% target for the more predictable
outcomes (CRP and HbA1c) but falls below target for triglycerides, the
least predictable outcome; at the subgroup level we observe localised
undercoverage (e.g., HbA1c for Mexican American participants),
illustrating the gap between the marginal guarantees of split conformal
prediction and the conditional coverage that clinical fairness
ultimately requires.
All code and the derived benchmark dataset are publicly available at
\url{https://github.com/felizzi/nhanes-accel-cardiometabolic-benchmark}.

\end{abstract}

\section{Introduction}

Structured tabular data underlies the majority of clinical decision-making,
spanning electronic health records, laboratory results, and wearable sensor
summaries~\citep{jiang2025survey}. Despite rapid progress in tabular
representation learning---from gradient-boosted trees to transformer
architectures and tabular foundation models---benchmark evaluations
predominantly use generic datasets that omit properties critical to
health applications: complex survey sampling designs, population
representativeness, clinically validated outcome measures, and
regulatory requirements for equitable performance across demographic
subgroups~\citep{mcelfresh2023neural}.

Wearable accelerometers offer a compelling avenue for population-scale
digital biomarker discovery. Physical activity measured continuously over
seven days captures behavioural patterns meaningfully associated with
cardiometabolic risk, yet translating raw activity counts into clinically
actionable predictions requires models that handle heterogeneous mixed-type
features, missing laboratory values, and---critically---must provide
reliable uncertainty estimates to support clinical deployment.

We make four contributions. \textbf{First}, we introduce the
\emph{NHANES Accelerometry Cardiometabolic Benchmark}, a population-representative
health tabular benchmark with properties absent from existing suites:
survey design weights, demographic oversampling of minority groups,
clinically validated fasting outcomes, and two-wave temporal structure.
\textbf{Second}, we conduct a rigorous comparison of methods spanning
the taxonomy of~\citet{jiang2025survey}---from specialised classical
models to general tabular foundation models---on three cardiometabolic
outcomes.
\textbf{Third}, we apply \emph{split conformal prediction}~\citep{angelopoulos2023gentle}
to all models, producing distribution-free prediction intervals with
finite-sample coverage guarantees.
\textbf{Fourth}, we evaluate whether conformal coverage holds equitably
across the sex and race/ethnicity subgroups that NHANES was designed
to represent, directly addressing the fairness gap identified
in~\citet{jiang2025survey}.

\section{Data and Benchmark}

\subsection{NHANES 2003--2006}

The National Health and Nutrition Examination Survey (NHANES) is a
stratified, multistage probability sample of the non-institutionalised
US civilian population, conducted by the Centers for Disease Control
and Prevention~\citep{nhanes2003}. We use examination cycles C
(2003--2004) and D (2005--2006), which included hip-worn ActiGraph
accelerometry for up to seven consecutive days.

\subsection{Analytic Sample}

Starting from participants with valid accelerometry data
($\geq$4 wear days, $\geq$600~min/day wear time), we applied the
following inclusion criteria: age 20--85 years; non-missing HbA1c,
triglycerides, and CRP; and fasting duration $\geq$8~hours prior to
blood draw (required for valid lipid measurements).
The final analytic sample comprised \textbf{N=1,381} participants.
Descriptive statistics are reported in Table~\ref{tab:descriptive}.

\begin{table}[h]
\caption{Analytic sample characteristics (N=1,381).}
\label{tab:descriptive}
\vskip 0.1in
\begin{center}
\begin{small}
\begin{tabular}{lccc}
\toprule
Variable & Mean & SD & Median \\
\midrule
Age (years)              & 52.7  & 18.6  & 53.0  \\
BMI (kg/m$^2$)           & 28.3  &  5.9  & 27.5  \\
HbA1c (\%)               &  5.6  &  0.9  &  5.4  \\
Triglycerides (mg/dL)    & 153.3 & 146.7 & 123.0 \\
CRP (mg/dL)              &  0.5  &  1.0  &  0.2  \\
TAC ($\times10^3$ counts/day) & 251.3 & 139.9 & 226.8 \\
MVPA (min/day)           & 20.6  & 22.6  & 13.1  \\
Sedentary (min/day)      & 1086.7 & 115.6 & 1093.0 \\
Energy (kcal/day)        & 2149.0 & 931.9 & 1990.5 \\
\midrule
Female (\%)              & \multicolumn{3}{c}{50.6\%} \\
NH White (\%)            & \multicolumn{3}{c}{57.1\%} \\
Mexican American (\%)    & \multicolumn{3}{c}{20.8\%} \\
NH Black (\%)            & \multicolumn{3}{c}{15.9\%} \\
Other (\%)               & \multicolumn{3}{c}{6.2\%}  \\
\bottomrule
\end{tabular}
\end{small}
\end{center}
\vskip -0.1in
\end{table}

\subsection{Features}

\textbf{Activity features} (6): total activity counts (TAC),
log-TAC (TLAC), sedentary time (ST), moderate-to-vigorous PA (MVPA),
light PA (LIPA), and wear time (WT), computed from minute-level
accelerometry using standard cutpoints~\citep{troiano2008physical}.

\textbf{Demographic and clinical covariates} (12): age, sex,
race/ethnicity, poverty-income ratio, BMI, systolic blood pressure,
smoking status, and total energy, carbohydrate, fat, protein, and
fibre intake from 24-hour dietary recall.

\textbf{Outcomes}: log-transformed HbA1c (\%), fasting triglycerides
(mg/dL), and CRP (mg/dL), each modelled separately.

\section{Methods}

\subsection{Experimental Setup}

We partitioned the analytic sample into train (60\%), calibration (20\%),
and test (20\%) sets using stratified random splitting (seed=42).
Numerical features were standardised using training-set statistics.
Categorical features (sex, race/ethnicity, smoking) were label-encoded.
All outcomes were log-transformed prior to modelling to address
right-skewed distributions; reported metrics are computed on the
log scale.

\subsection{Models}

We evaluated three models spanning the specialised-to-general
taxonomy of~\citet{jiang2025survey}:

\textbf{Ridge Regression} ($\alpha$=1.0): a linear specialised baseline
providing an interpretable lower bound on predictive performance.

\textbf{XGBoost}~\citep{chen2016xgboost}: a gradient-boosted tree
ensemble representing the current state of practice for clinical
tabular prediction. We used 500 estimators with learning rate 0.05,
max depth 6, and early stopping on the calibration set.

\textbf{TabPFN~v2}~\citep{hollmann2025tabpfn} is a tabular foundation
model representing the \emph{general} tier of the taxonomy
of~\citet{jiang2025survey} --- it requires no task-specific training,
applying a learned prior directly to downstream tasks without fine-tuning.
TabPFN~v2 is pre-trained via \emph{in-context learning} on millions of
synthetically generated datasets constructed from structural causal models
and Bayesian neural networks, enabling the model to perform approximate
Bayesian inference over tabular inputs at test time.
Concretely, given a training set $\mathcal{D}_\text{train}$ and a test
instance $x^*$, TabPFN~v2 computes:
\begin{equation}
    \hat{y}^* = g_\Theta(x^* \mid \mathcal{D}_\text{train})
\end{equation}
where $g_\Theta$ is a transformer whose weights $\Theta$ are fixed after
pre-training --- the training set is passed directly as context, and
predictions are produced in a single forward pass without gradient updates.
This \emph{in-context} mechanism allows TabPFN~v2 to implicitly perform
model selection and uncertainty estimation within its forward pass,
which we hypothesise accounts for its advantage over task-specifically
trained baselines on our modest sample size (N$_\text{train}$=828).
Following~\citet{hollmann2025tabpfn}, training instances are passed
directly as in-context examples, with subsampling to 1,000 examples
applied only when the training set exceeds this size; here
N$_\text{train}$=828 $<$ 1{,}000, so all training examples were used as
context without subsampling. We used CPU inference throughout.

\subsection{Conformal Prediction}

We applied \emph{split conformal prediction}~\citep{angelopoulos2023gentle}
to each model using the calibration set. For each model and outcome,
we computed the non-conformity score as the absolute residual
$|y_i - \hat{y}_i|$ on the calibration set, and formed prediction
intervals at miscoverage level $\alpha$=0.10 (targeting 90\% coverage):
\begin{equation}
    \hat{C}(x) = [\hat{y} - \hat{q}, \; \hat{y} + \hat{q}]
\end{equation}
where $\hat{q}$ is the $\lceil(1-\alpha)(1+1/n_\text{cal})\rceil$
quantile of calibration residuals. Split conformal prediction
provides a finite-sample marginal coverage guarantee without
distributional assumptions~\citep{angelopoulos2023gentle}.

\subsection{Subgroup Coverage Analysis}

We evaluated whether conformal coverage held equitably across
demographic subgroups defined by sex (male/female) and race/ethnicity
(Mexican American, Other Hispanic, Non-Hispanic White, Non-Hispanic
Black, Other), using the NHANES oversampling design that ensures
adequate representation of minority groups.

\section{Results}

\subsection{Predictive Performance}

Table~\ref{tab:results} reports R$^2$, mean absolute error (MAE,
log scale), and empirical conformal coverage for all model--outcome
combinations.

\begin{table}[h]
\caption{Model comparison on test set (N=276). Coverage target = 0.90.
MAE reported on log scale.}
\label{tab:results}
\vskip 0.1in
\begin{center}
\begin{small}
\begin{tabular}{llccc}
\toprule
Outcome & Model & R$^2$ & MAE & Cov. \\
\midrule
\multirow{3}{*}{HbA1c}
  & Ridge     & 0.133 & 0.073 & 0.910 \\
  & XGBoost   & 0.139 & 0.073 & 0.881 \\
  & TabPFN v2 & \textbf{0.156} & \textbf{0.070} & 0.884 \\
\midrule
\multirow{3}{*}{Triglycerides}
  & Ridge     & $-$0.002 & 0.474 & 0.863 \\
  & XGBoost   & $-$0.007 & 0.470 & 0.823 \\
  & TabPFN v2 & \textbf{0.048} & \textbf{0.453} & 0.819 \\
\midrule
\multirow{3}{*}{CRP}
  & Ridge     & 0.339 & 0.787 & 0.917 \\
  & XGBoost   & 0.313 & 0.809 & 0.913 \\
  & TabPFN v2 & \textbf{0.383} & \textbf{0.767} & \textbf{0.939} \\
\bottomrule
\end{tabular}
\end{small}
\end{center}
\vskip -0.1in
\end{table}

TabPFN~v2 achieves the best R$^2$ on all three outcomes, with the
largest margin on CRP (R$^2$=0.383 vs.\ 0.339 for ridge).
The advantage of TabPFN~v2 is most pronounced for CRP
($\Delta$R$^2$=+0.044 over ridge, +0.070 over XGBoost),
consistent with its in-context Bayesian prior capturing
non-linear interactions between activity, BMI, and dietary
features that are poorly approximated by linear models and
insufficiently regularised in gradient-boosted trees at this
sample size.
CRP is the most predictable outcome across all models, consistent
with the strong behavioural determinants of systemic inflammation.
Triglycerides show near-zero or negative R$^2$ for all models,
reflecting the dominant role of acute dietary intake and genetics
neither captured in our feature set nor available in NHANES
public-use files~\citep{teslovich2010biological}.
HbA1c is moderately predictable (R$^2$=0.133--0.156), consistent
with published estimates using lifestyle variables in general
population samples~\citep{selvin2004glycemic}.

Conformal coverage is closest to the 90\% target for ridge regression
(HbA1c: 0.910, CRP: 0.917) and TabPFN~v2 on CRP (0.939). XGBoost
shows slight undercoverage on triglycerides (0.823), reflecting
greater overconfidence in its residual distribution.

\subsection{Subgroup Coverage Analysis}

\begin{figure}[h]
\begin{center}
\includegraphics[width=\columnwidth]{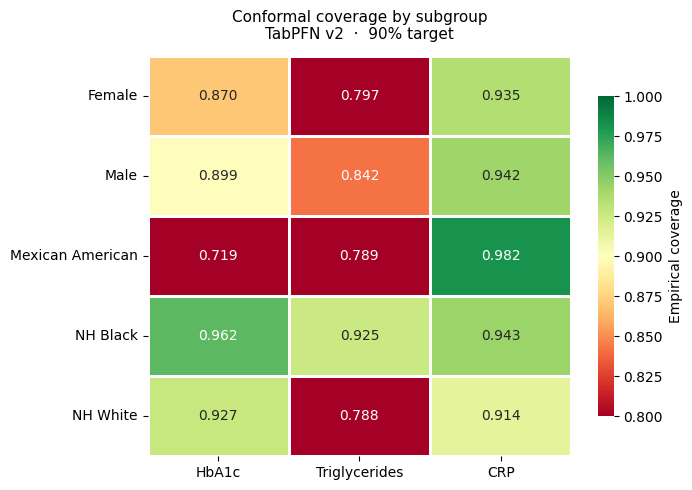}
\end{center}
\caption{Empirical conformal coverage (90\% target) by demographic
subgroup for TabPFN~v2. Colour indicates coverage level:
green $\geq$0.90, red $<$0.90.}
\label{fig:subgroup}
\end{figure}

Figure~\ref{fig:subgroup} shows empirical conformal coverage by demographic subgroup for TabPFN~v2. Coverage is broadly consistent across sex and race/ethnicity subgroups for CRP, with all groups meeting or exceeding the 90\% target. However, notable undercoverage is observed for Mexican American participants on HbA1c (0.719), while other groups achieve $\geq 0.87$ coverage on this outcome. Undercoverage is also observed for triglycerides across all subgroups, consistent with the poor overall predictability of this outcome. Notably, coverage for Non-Hispanic Black participants meets or exceeds the 90\% target on both HbA1c and CRP, indicating that the marginal intervals remain valid for this group despite its higher prevalence of cardiometabolic risk.

\section{Discussion}

\paragraph{TabPFN v2 as a strong baseline for small health datasets.}
Our results show that the tabular foundation model TabPFN~v2
achieves the best point performance among the evaluated baselines on
all three outcomes without any task-specific training, although on this
single split the margins over ridge and XGBoost are modest and should be
confirmed with repeated-split evaluation. This is consistent with findings
in~\citet{hollmann2025tabpfn} showing that TabPFN~v2 achieves
state-of-the-art performance on small tabular datasets
(N$<$10,000), and extends this finding to a clinically structured,
population-representative health benchmark.

\paragraph{Why TabPFN v2 outperforms task-specific models.}
The superior performance of TabPFN~v2 on a dataset of N=1,381
is consistent with its design objective: the model's pre-training
on synthetic datasets spanning diverse causal structures effectively
encodes a broad prior over tabular relationships, which acts as strong
regularisation when task-specific training data is scarce.
Classical models such as ridge regression and XGBoost must
estimate all structure from the available 828 training examples,
while TabPFN~v2 leverages knowledge distilled from millions of
synthetic datasets to inform its predictions.
This finding extends results from~\citet{hollmann2025tabpfn} ---
which demonstrated TabPFN~v2's advantage on generic small tabular
benchmarks --- to a clinically structured, population-representative
health dataset, suggesting that tabular foundation models may be
particularly well-suited to health applications where data collection
is expensive and sample sizes are inherently limited.

\paragraph{Outcome-specific predictability.}
The strong contrast between CRP (R$^2$$\approx$0.38) and
triglycerides (R$^2$$\approx$0.00) highlights a key challenge for
digital biomarker discovery: not all cardiometabolic outcomes are
equally predictable from lifestyle and activity data. Triglycerides
are dominated by genetic factors~\citep{teslovich2010biological} and
acute dietary intake, neither of which is captured by 7-day
accelerometry or single 24-hour dietary recall. Future work
incorporating genetic data or repeated dietary assessments may
improve triglyceride prediction substantially.

\paragraph{Marginal versus conditional coverage.}
Split conformal prediction provides valid \emph{marginal} coverage
without distributional assumptions, making it an attractive starting
point for the heterogeneous population structure of NHANES. Marginal
validity, however, does not imply conditional validity: the guarantee
holds on average over the test distribution and can be violated within
individual subgroups. Our analysis bears this out --- coverage is close
to nominal for most groups on CRP and HbA1c, but drops to 0.719 for
Mexican American participants on HbA1c, and triglyceride intervals
undercover across the board. Marginal conformal methods alone are
therefore insufficient to certify demographic fairness. Group-conditional
(Mondrian) conformal calibration, which forms separate calibration
quantiles per subgroup, is a natural remedy, although exact conditional
coverage is generally unattainable in finite samples without further
distributional assumptions~\citep{angelopoulos2023gentle}; we leave a
systematic comparison to future work.

\paragraph{Limitations.}
Our evaluation uses a single train/calibration/test split (seed=42).
While sufficient to establish the benchmark and to report empirical
coverage, the small test set (N=276) and the narrow R$^2$ gaps between
models (e.g., 0.133 vs.\ 0.156 for HbA1c) mean that point-estimate
differences should be read as indicative rather than as statistically
established rankings; a fixed split on a small sample can also induce
calibration--test distribution shift, which partly explains the residual
undercoverage we observe. The benchmark is also limited in breadth:
we compare three models and three cardiometabolic markers (HbA1c,
triglycerides, CRP), leaving a fuller panel (e.g., fasting glucose,
HDL/LDL cholesterol) and additional learners for future iterations.
The sample is predominantly Non-Hispanic White (57.1\%),
reflecting the fasting requirement which differentially excluded
participants from minority groups, and results may not generalise
equally across all populations.
The 2003--2006 protocol uses hip-worn, predominantly uniaxial ActiGraph
devices, which differ substantially from the wrist-worn triaxial sensors
in contemporary consumer and research wearables; the cutpoint-based
activity features and the trained models may transfer poorly to modern
data streams, and we make no claim of device-level generalisation.
The single 24-hour dietary recall
introduces measurement error in dietary covariates. NHANES does not
include genetic data, precluding adjustment for heritable determinants
of cardiometabolic risk. Sedentary time as computed includes
non-wear periods and should be interpreted with caution.

\paragraph{Future work.}
Several directions follow directly from the limitations above.
\emph{Robustness}: repeating the full pipeline over multiple random
splits and reporting mean$\pm$std with paired tests on per-seed
differences would put the model rankings on firmer statistical footing.
\emph{Calibration}: replacing split conformal with cross-conformal or
jackknife+~\citep{barber2021predictive} calibration would use the full
sample for both fitting and calibration, reducing the sensitivity of
coverage to a single calibration draw, and group-conditional (Mondrian)
calibration would target subgroup-level validity directly.
\emph{Ablations}: adding survey-weighted ridge/GBM baselines (consistent
with the population-representative framing) and a demographics- or
BMI-only baseline for CRP would isolate the incremental signal
contributed by accelerometry beyond established anthropometric
predictors.
\emph{Scope}: incorporating NHANES 2011--2014 wrist
accelerometry for larger samples and richer signal; broadening the
outcome panel to fasting glucose and HDL/LDL cholesterol for a more
comprehensive cardiometabolic profile; evaluating
FT-Transformer~\citep{gorishniy2021revisiting} at sufficient sample
sizes (N$>$5,000) where transformer architectures are expected to
realise their representational advantages~\citep{jiang2025survey};
and extending conformal prediction to survival outcomes using
NHANES linked mortality data.

\section{Conclusion}

We introduced the NHANES Accelerometry Cardiometabolic Benchmark as a
population-representative health tabular benchmark and showed that the
tabular foundation model TabPFN~v2 achieves the strongest point
predictions among the evaluated baselines on all three cardiometabolic
outcomes, with the caveat that these comparisons rest on a single data
split and modest margins. Split conformal intervals attain
close-to-nominal \emph{marginal} coverage for the more predictable
outcomes, while subgroup analysis exposes localised undercoverage that
marginal guarantees do not preclude --- motivating conditional and
group-aware calibration as the natural next step. Our benchmark and code
provide a reproducible foundation for future comparisons of tabular
learning methods on clinically structured health data.


\bibliography{references}
\bibliographystyle{icml2026}

\end{document}